%% file: acl_latex.tex
\pdfoutput=1

\documentclass[11pt]{article}

\usepackage[]{acl}

\usepackage{times}
\usepackage{latexsym}
\usepackage{graphicx}
\usepackage{adjustbox}
\usepackage{array}
\newcolumntype{L}{>{\centering\arraybackslash}m{3cm}}
\newcolumntype{M}[1]{>{\begin{varwidth}[t]{#1}}l<{\end{varwidth}}}
\usepackage[T1]{fontenc}

\usepackage[utf8]{inputenc}

\usepackage{microtype}

\title{ Massively Multilingual Language Models for Cross Lingual Fact Extraction from Low Resource Indian Languages}

\author{Bhavyajeet Singh\textsuperscript{*}, Pavan Kandru\textsuperscript{*}, Anubhav Sharma \and Vasudeva Varma \\
        Information Retrieval and Extraction Lab, IIIT Hyderabad, India \\  \{bhavyajeet.singh, siri.venkata, anubhav.sharma\}@research.iiit.ac.in \\ vv@iiit.ac.in}

\begin{document}
\maketitle
\let\thefootnote\relax\footnotetext{\textsuperscript{*} Equal contribution}

\input{sections/0_abstract}

\input{sections/1_introduction}

\input{sections/2_related_work}

\input{sections/3_dataset}

\input{sections/4_methodology}

\input{sections/5_results}

\input{sections/6_conclusion}

\bibliography{anthology,custom}
\bibliographystyle{acl_natbib}

\newpage
\clearpage

\appendix
\input{sections/7_appendix}

\end{document}

%% file: sections/0_abstract.tex
\begin{abstract}
\vspace{-2mm}

Massive knowledge graphs like Wikidata attempt to capture world knowledge about multiple entities. 
Recent approaches concentrate on automatically enriching these KGs from text. However a lot of information present in the form of natural text in low resource languages is often missed out. 
Cross Lingual Information Extraction aims at extracting factual information in the form of English triples from low resource Indian Language text. 
Despite its massive potential, progress made on this task is lagging when compared to Monolingual Information Extraction. In this paper, we propose the task of Cross Lingual Fact Extraction(CLFE) from text and devise an end-to-end generative approach for the same which achieves an overall F1 score of 77.46.

\end{abstract}
\vspace{-2mm}

%% file: sections/1_introduction.tex
\section{Introduction}
\vspace{-2mm}

Knowledge graphs are large structured sources of information about the world. Recently, a lot of attention is being put in finding ways to automatically build or enrich extensive knowledge graphs (KGs) \cite{kgfromtext1}, \cite{Zou_2020}. Wikidata \cite{Wikidata_creation} is one of the largest publicly available knowledge graphs which has over 99 million entities. Knowledge graphs such as Wikidata have been extensively used for multiple applications like text generation \cite{text_from_kg}, question answering \cite{QA_from_kg_1}, \cite{QA_from_kg} etc. 

A knowledge graph is composed of multiple facts linked together. A fact is often represented as a triplet which consists of two entities and a semantic relation between them. This information can be encoded as a triple $<h,r,t>$ where $h$ is the subject entity, $r$ is the relation and $t$ represents the tail entity.

Fact extraction refers to the task of extracting structured factual information from natural language text \cite{book}.
Previously there has been extensive work regarding the task of monolingual fact extraction, especially in English \cite{copyMTL} \cite{jointIE}, however not much attention has been given to the task of cross-lingual fact extraction. 

\begin{figure}[h]
    \centering
    \includegraphics[width=0.5\textwidth]{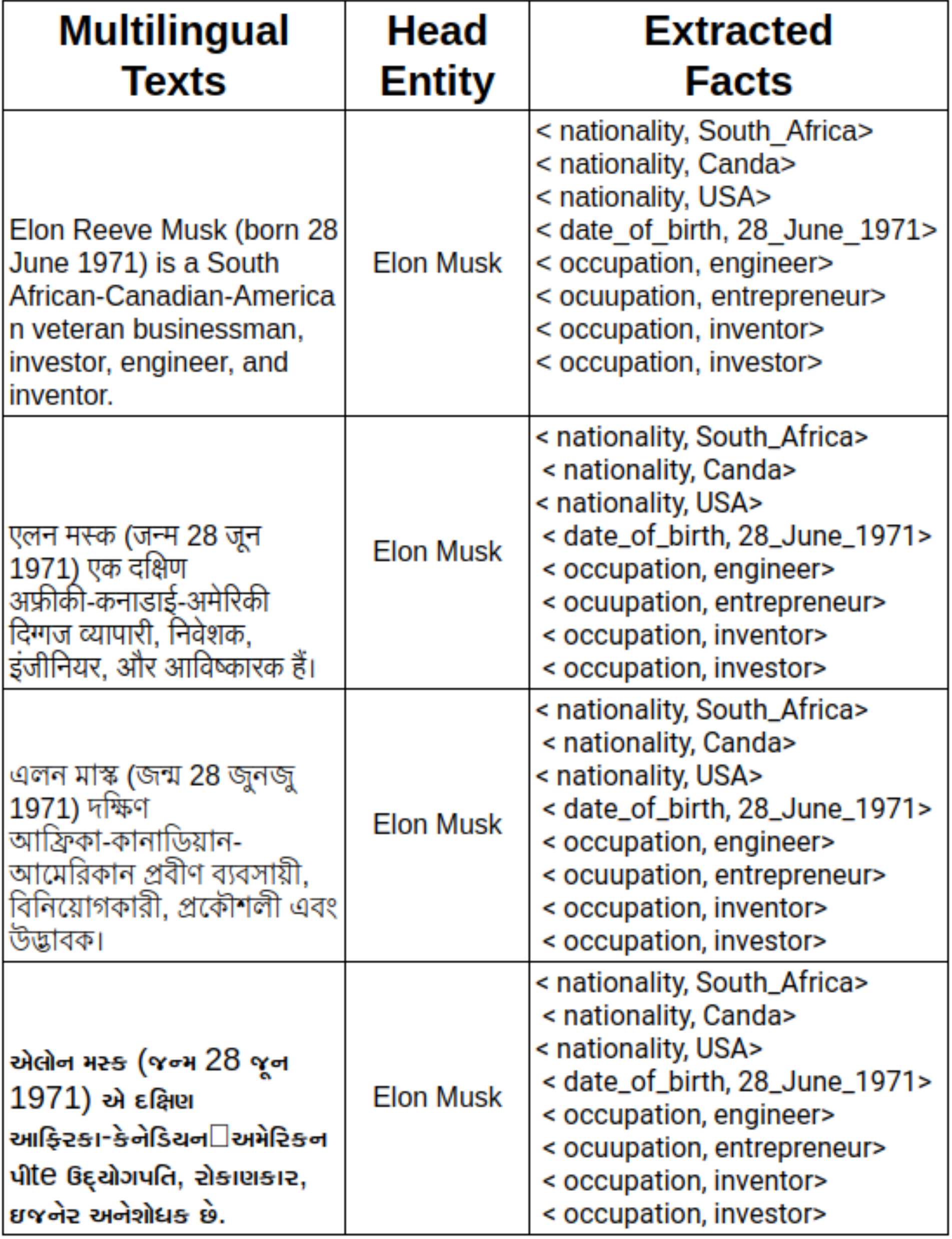}
    \caption{Example Inputs and outputs of CLFE task. Text from any language along with entity of interest(head entity) is provided as input to extract English Facts(relation and tail entity pairs). The same sentence may or may not be present in all languages.}
    \label{fig:example}
\vspace{-2mm}
\end{figure}

In this paper we propose an important task of multi-lingual and cross-lingual fact to text extraction(CLFE) for 7 Low Resource(LR) Indian Languages and English. The task aims at directly extracting English triples from 8 different languages. We also propose strong baselines and approaches\footnote{code available at \href{https://github.com/bhavyajeet/CLFE}{https://github.com/bhavyajeet/CLFE}} for this task which produce results comparable to existing mono-lingual state-of-the-art fact extraction pipelines and significantly better than other previous cross lingual attempts at fact extraction \cite{chinese_cross}. Our work enables utilisation of factual knowledge present in Indic texts in order to increase the coverage of existing knowledge graphs. This would further help in multiple downstream tasks like fact verification, text generation etc. To the best of our knowledge, this is the first attempt at multilingual and cross-lingual fact extraction from LR Indian Languages. 
Figure \ref{fig:example} shows multiple examples of the input and output for CLFE task. 

Overall, we make the following contributions.
(1) Propose the problem of cross-lingual and multilingual fact extraction for LR Indian languages. (2) An end-to-end generative approach  for extracting subject centric factual information from LR Indian language text, which shows significant improvements over classification based pipelines. (3) We train multiple multi-lingual CLFE models which lead to an overall F1 score of 77.46 .


%% file: sections/2_related_work.tex
\section{Related work}
\vspace{-2mm}

Extracting structured information from free form text is a problem well worked upon. T-REx \cite{Trex} uses entity linking, co-reference resolution and string match based linking pipelines to perform fact linking between DBPedia \cite{dbpedia} abstracts and Wikidata \cite{Wikidata_creation} triples. REFCOG \cite{Refcog} works in a cross lingual space
to link the facts and outperforms the existing pipeline based approaches like \cite{Trex}. But these approaches are limited in their application since they perform fact linking and need a fact set as input. 

OpenIE\cite{openie} tackles this issue by leveraging linguistic structure for open domain information extraction. While the predecessor open domain IE systems like Ollie\cite{ollie}  use large set of patterns with broad coverage to extract facts, OpenIE uses a small set of patterns which works well on canonically structured sentences. However, these open domain information extractors produce facts that have long and over-specific relations which can not be used to construct KGs.

\cite{frusteasy}; \cite{setprediction} approach the information extraction problem by jointly extracting entities and their relation from input text using neural models without referring to any repository of facts. Although these works produce systems which can extract open information from text in the WebNLG dataset \cite{webnlg}, they are monolingual and are limited to knowledge extraction in a single language. Various existing well performing relation extraction models like \cite{pfn}, \cite{set_pred} rely partially on exact match of entities in the source text, which makes it harder to adapt them for the CLFE task. 

Cross Lingual fact extraction i.e extracting facts from source text of different languages didn't receive as much attention as Monolingual Fact extraction did. Although \cite{zhang} worked on this task, with just a single language, the highest reported f1 is 33.67. Moreover, Fact extraction from low resource languages like Indic Languages hasn't been attempted. In this work, we attempt to reduce these gaps in Information extraction by proposing systems for Cross Lingual Subject Centric Fact Extraction in low resource Indic Languages. 

\begin{figure*}[ht]
    \centering
    \includegraphics[width=0.95\textwidth]{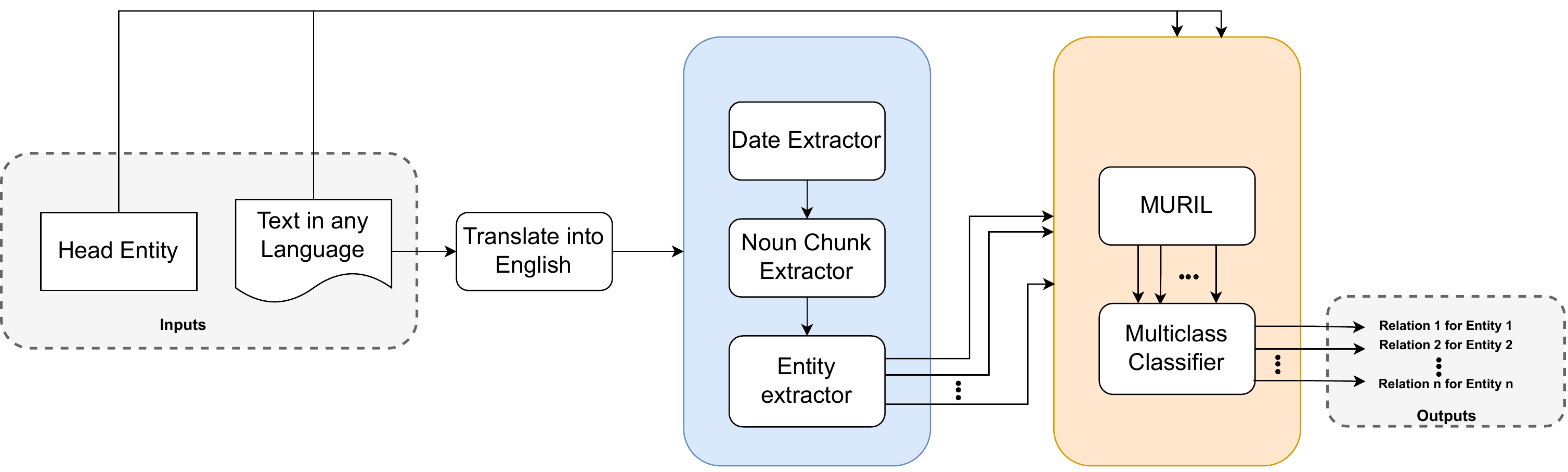}
    \caption{Pipeline Architecture for CLFE}
    \label{fig:pipeline}
\end{figure*}

\begin{figure*}[ht]
    \centering
    \includegraphics[width=0.95\textwidth]{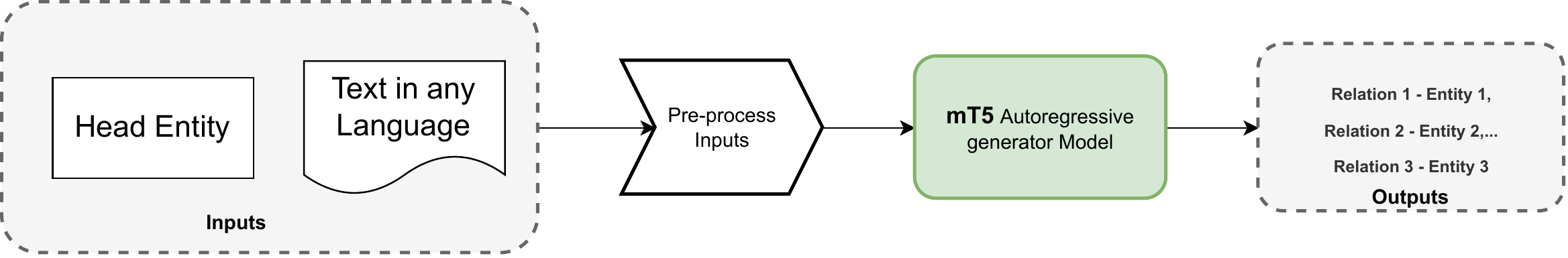}
    \caption{End to end architecture for CLFE}
    \label{fig:e2e}
\end{figure*}

%% file: sections/3_dataset.tex
\section{Dataset}
\vspace{-2mm}

For the task of CLFE we leverage the XAlign dataset \cite{Xalign_cite}. The dataset contains 0.45M pairs across 8 languages, of which 5402 pairs have
been manually annotated. The manually annotated part of the dataset was used as the golden test set. The sentences in XAlign come from the Wikipedia articles, about entities belonging to the human class, written in Indian languages.

The extensively cross lingual and multi lingual nature of the XAlign dataset is ideal for the proposed task. Though originally designed for the task of cross lingual data to text generation, the XAlign dataset can be leveraged for CFLE as well. 
However the dataset poses certain challenges.  If we were to consider each relation as a class (for classification based approaches), the dataset is highly imbalanced. Out of approximately 367 unique relations(classes), the most frequent class alone has a frequency of 27 \% and top 20 classes contribute to 90\% of the data. The data contains an average of 2.02 facts aligned per sentence. 

Along with this, another challenging aspect of the dataset is that it is partially aligned. While the sentences in the test set have complete coverage in the aligned facts, the entire information present in the sentences from the train set is not covered by the aligned facts. This attribute of the dataset, can potentially penalise the model even for the generation of correct facts during the training time. Thus impacting recall scores during the test time. More details in Appendix section \ref{sec:adddata}

\begin{table*}[ht]
\centering
\tabcolsep=0.12cm
\begin{adjustbox}{max width=\textwidth}

\begin{tabular}{Lccccccccccc}
\hline
\textbf{} & \textbf{te} & \textbf{bn} & \textbf{ta} & \textbf{gu} & \textbf{mr} & \textbf{en} & \textbf{hi} & \textbf{kn} & \multicolumn{3}{c}{\textbf{All languages}} \\
\hline
 &
  F1 &
  F1 &
  F1 &
  F1 &
  F1 &
  F1 &
  F1 &
  F1 &
  P &
  R &
  F1 \\
\hline
\textbf{Classification with GT Tails} &
  69.19 &
  67.50 &
  89.44 &
  85.74 &
  51.38 &
  72.87 &
  87.10 &
  79.74 &
  79.04 &
  77.93 &
  75.37 \\
 \hline
\textbf{TERC} &
  43.66 &
  41.96 &
  52.19 &
  40.30 &
  44.59 &
  50.80 &
  50.46 &
  42.57 &
  40.45 &
  53.71 &
  46.15 \\
 \hline
\textbf{E2E Cross-lingual Generative Model} &
  71.82 &
  75.56 &
  82.82 &
  \textbf{72.36} &
  \textbf{77.79} &
  76.28 &
  \textbf{86.62} &
  68.04 &
  74.09 &
  \textbf{81.15} &
  \textbf{77.46} \\
 \hline
\textbf{E2E generation w script unification} &
 \textbf{72.51} &
  75.38 &
  \textbf{85.21} &
  72.04 &
  77.19 &
  74.56 &
  83.44 &
  \textbf{70.46} &
  78.49 &
  76.15 &
  77.29 \\
  \hline
\textbf{Bilingual Models} &
  70.94 &
  \textbf{78.01} &
  83.71 &
  67.84 &
  71.91 &
  \textbf{76.64} &
  86.49 &
  63.19 &
  \textbf{79.79} &
  71.63 &
  75.49 \\
  \hline
\end{tabular}
\end{adjustbox}
\caption{Precision, recall and F1 scores of various methods applied on all languages in the Test set. Note that "Classification with GT Tails" uses tails from ground truth as input for the Relation Prediction model and hence does not represent a complete pipeline}
\label{tab:results}
\end{table*}

%% file: sections/4_methodology.tex
\section{Methodology}
\label{sec:4}
\vspace{-1mm}

We propose two approaches for the CLFE task. The first approach is a classification based approach which extracts tails first and then predicts the relation. Second approach is a generative one that does both of these task in one shot.

\subsection{Tail Extraction and Relation Classification(TERC)}
\label{sec:terc}
The TERC pipeline (Figure \ref{fig:pipeline}) consists of two steps. The first step is to extract tails of facts from the source language text. To do this we use IndicTrans \cite{samanantar} translation and convert input text to English language. After this we extract any dates present in the text and normalize them in to the same format. We also replace them in the original text with a dummy token to preclude dates from participating in other entities. Since every tail entity can only be a noun or proper noun, we use spaCy \cite{honnibal_spacy_2018} noun chunk extractor to extract all the noun chunks from which tail entities are selected as follows.
\vspace{-2mm}
\begin{itemize}
\item  Entities that match with head are removed.
Since we are only interested about tails at this stage of the pipeline we remove any entities that have high lexical overlap with head.
\vspace{-3.25mm}
\item  All noun chunks with pronoun roots are removed to filter pronouns. 
Tails present in the data are never pronouns so we prune out all the recognized phrases which have pronoun heads.
\vspace{-3.25mm}
\item  Continuous spans of tokens with ADJ and PROPN PoS tags are selected as individual entities.
Tails are multi word entities and may contain adjectives within their span, so we use PoS tags to get maximal spans for every detected proper nouns.
\vspace{-3.25mm}
\item  Root of the noun chunk is selected as a separate entity if its PoS tag is NOUN.
\end{itemize}
\vspace{-1.5mm}
Next step is to  predict a relation for each of these tails. To do this we use pretrained MuRIL \cite{muril} to generate a joint representation of head entity, tail entity and source language input text. This representation is fed as input to a classifier which predicts the relation between the head and the tail entities in the input. The classifier is trained on the training set to predict the relation, given a sentence and a <head, tail> pair, by considering the tails from ground truth as input. In order to tackle the class imbalance, we use \textbf{inverse log of class distribution} as weights in loss-function which performs better than standard inverse class distribution as well as unweighted loss. 

While evaluating the performance of the pipeline architecture, tails extracted from translated input text, are aligned with ground truth tails. The details of this alignment are described in \ref{sec:tailalignment} of the Appendix.
Predictions are made for the aligned tails and evaluation metrics are calculated on the same.

\subsection{End to End Generative extraction}

We also propose an end to end approach (Figure \ref{fig:e2e}) to the fact extraction problem which can jointly extract tails as well as their relations with the head entity. Previous work in the domain of monolingual fact extraction has shown that a model which jointly performs the tail and relation extraction is more likely to perform better than a disjoint approach \cite{jointIE}. Advantage of this approach over the pipeline approach mentioned above is that there is a two way interaction between tail extraction and relation prediction which improves performance of both the tasks as they are not independent of each other.

We pose this problem as a text-to-text task and use the mT5 \cite{mt5} auto-regressive seq-2-seq model to generate relations and tails, when head entity and input text are given as inputs. We  use cross entropy loss to train this model. Using a generative model allows for a more generalizable and open information extraction i.e set of relations and tails are not restricted.

We experiment with 3 variations of this pipeline. In all these variations, the facts are linearised as the target text by concatenating the head and tail joined by special tokens. Thus for a given sentence $S$, if the corresponding $i$ facts are ${[h,r_1,t_1], [h,r_2,t_2] .... [h,r_i,t_i]}$, the target text would be $<R> r_1 <T> t_1 <R> r_2 <T> t_2 .... <R> r_i  <T> t_i$. 

The first variation is fine-tuning the pretrained mt5 model for the fact extraction task over all languages. For the second experiment, we use script unification where we transliterate the input text of all languages except English to the Devanagari script. The idea is that the unified script input helps the model's training due to a high overlap in the vocabulary accross multiple Indian languages. In our third variation, we train multiple bi-lingual fact extraction models, one for each language. The implementation details  of regarding these models and TERC(\ref{sec:terc}) are in the Appendix \ref{sec:impl}.

%% file: sections/5_results.tex
\section{Results and Discussion}
\vspace{-2mm}

Table \ref{tab:results} summarizes the results of the multiple fact extraction approaches mentioned in section \ref{sec:4}. 

It can be observed that the open ended approach performs the best in terms of F1 score while also providing complete flexibility regarding the possible entities and relations. Another observation is that the strategy where we train separate bilingual models, works better than the combined model for just two languages, English and Bengali. This is explained by the fact that these are the two most frequent languages for our dataset, which together constitute 54.44 \% of our training data.  Thus, multilingual training proves to be useful over all, because of the shared learning across Indian languages. 
We also observe that script unification (transliterating input scripts to Devanagari), specifically benefits all the Dravidian languages (te, ta, kn) of our dataset. 

 It should be noted that the actual performance of the model might be better than what the numbers show. The reason for this is that currently we adhere to strict evaluation schemes where a word match between the predicted and the actual tail is necessary in order to determine the prediction as correct. However, this misses out on cases where the predicted and the ground truth tails are completely synonymous. An example of this is the case where the model predicts the occupation as 'writer', whereas the GT label has it as 'author'.  

%% file: sections/6_conclusion.tex
\section{Conclusion and Future work}
\vspace{-2mm}

In this work, we introduce the task of multilingual and cross-lingual fact extraction over English and seven other LR Indic languages. We conclude that though script-unification helps certain languages, a single multilingual end-to-end generative pipeline performs better with overall F1 score of 77.46. This work paves the path for upcoming research in methods of extracting knowledge from LR Indic language text. In future, we plan to explore approaches that make specific effort to tackle the partially aligned nature of the dataset in order to achieve further improvements.

%% file: sections/7_appendix.tex
\section{Example Appendix}
\label{sec:appendix}

\subsection{Tail Entity Alignment}
\label{sec:tailalignment}
Entities extracted from the translated texts are aligned with the gold truth tail entities in order to measure performance on test set. By alignment we mean that we assign one ground truth tail entity to each extracted entity without repetition. Some extracted entities which do not have any overlap with ground truth are ignored. Some ground truth entities might not be assigned any of the extracted entities leading to lower recall. Assignment is done based on a similarity score and a threshold. Similarity score between two entities is calculated as the sum of cosine similarities of GloVe vectors and intersection over union of terms. With a threshold of 0.7 we achieved a precision of 0.54 and a recall of 0.77.

\subsection{Additional Dataset Statistics}
\label{sec:adddata}
\begin{figure}[!htbp]
    \centering
    \includegraphics[width=0.5\textwidth]{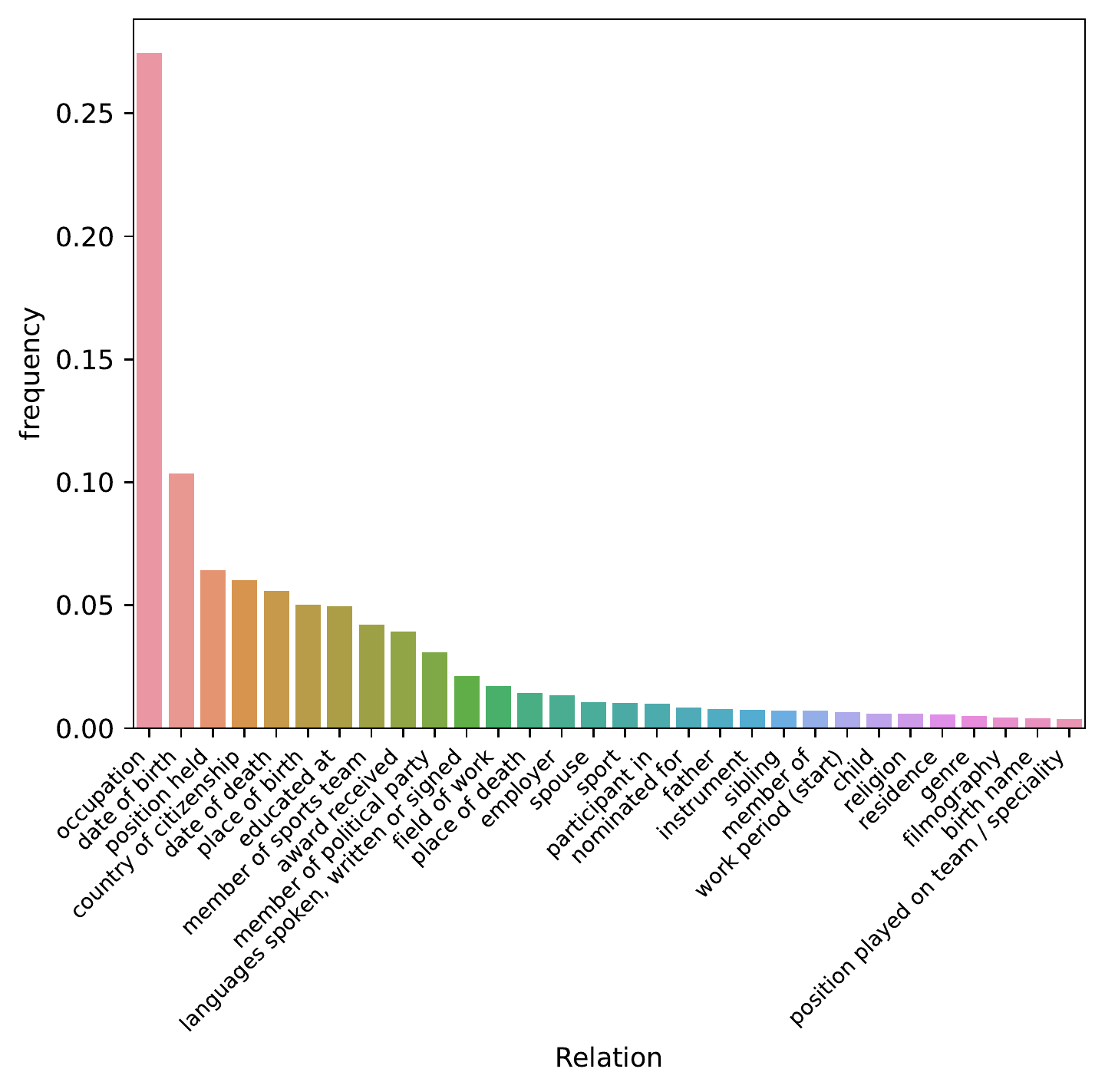}
    \caption{Distribution of Top 30 most frequent relations in the dataset}
    \label{fig:distribution}
\end{figure}

\begin{figure}[!htbp]
    \centering
    \includegraphics[width=0.5\textwidth]{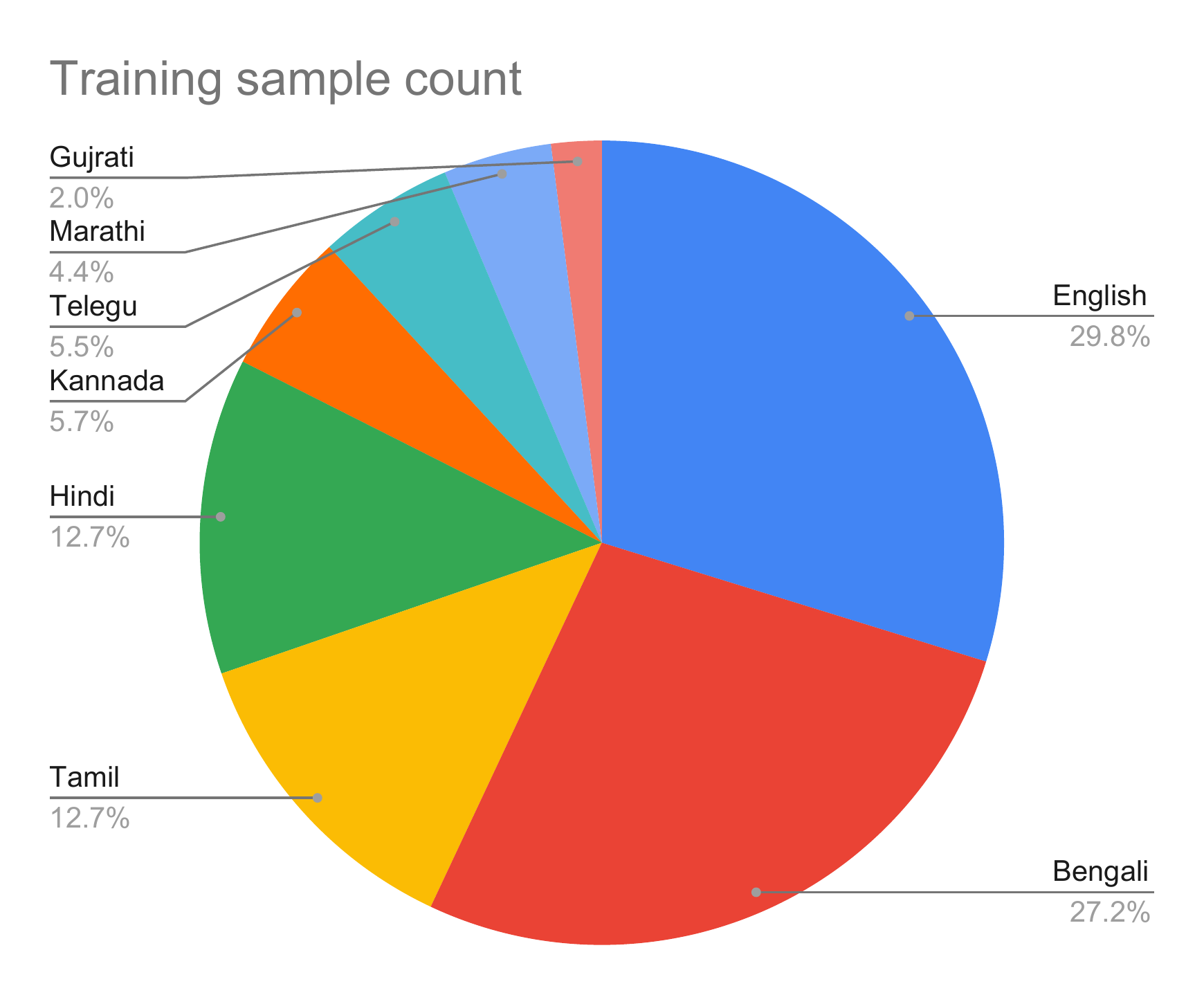}
    \caption{Distribution of the 8 languages in the training set}
    \label{fig:datasample}
\end{figure}

Figure \ref{fig:distribution} shows the distribution of Top 30 most frequent relations in the dataset. Figure \ref{fig:datasample} depicts the share of each of the languages in the dataset. As it can be seen, the dataset is highly imbalanced both in terms of relations and languages. 

\subsection{Implementation Details}
\label{sec:impl}
Both Two-Phase and E2E generative architectures are trained on NVIDIA GeForce RTX 2080 Ti graphic cards. For the Two-Phase approach, the only block that needed training is relation prediction. MURIL encoder model from google which has 12 encoding layers and output dimension 768 is the base of the classifier. 12th layer of MURIL along with the layers in the feed forward network are unfrozen during the training phase. Adam optimizer is used with initial learning rate of 1e-4 and step scheduling with step size 2 and gamma 0.3. Batches of 16 facts are trained to optimize Cross Entropy Loss. Inverse log frequency of classes is used as weights for cross entropy loss to counteract the imbalance in the dataset. Training relation prediction 5 hours on 1 GPU card.

For the Generative approach, we used the pretrain mT5 model and finetune it for 5 epochs for all experiments. The learning rate is 0.001 with a weight decay of 0.01. The dropout rate is set to 0.1 in order to prevent over fitting on the training data. We use the Adafactor optimizer to optimise the Cross Entropy Loss during generation. 

\subsection{Analysis of OpenIE extraction}
\begin{figure}[!htbp]
    \centering
    \includegraphics[width=1.1\textwidth]{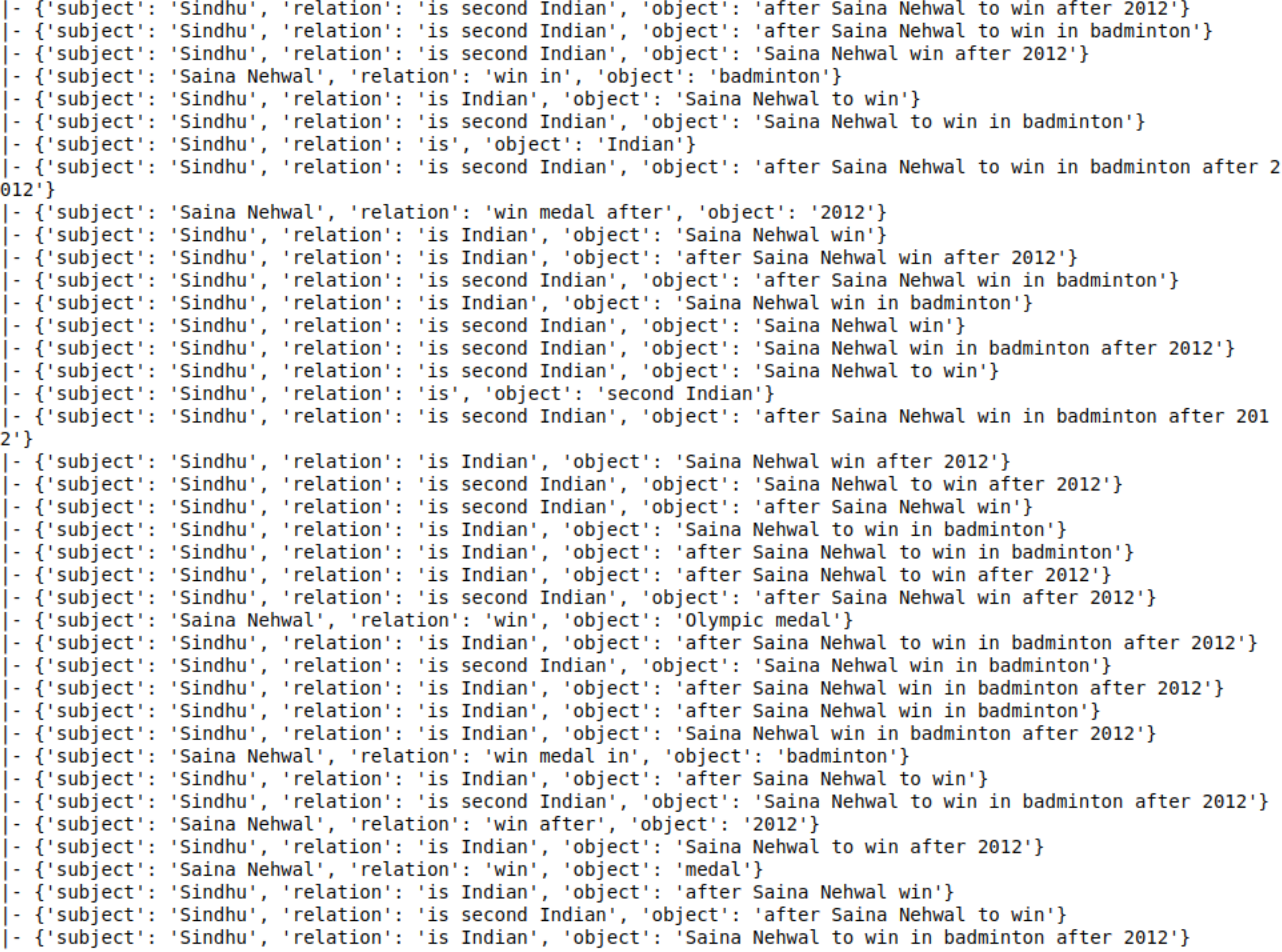}
    \caption{Examples of output from OpenIE}
    \label{fig:openie}
\end{figure}

We tried openIE extractor from stanford to extract from english translated versions of texts from other languages. Even after discounting translation losses facts extracted from openIE were not useful because of overly specific relations and entities. Figure \ref{fig:openie} is an example for the source sentence "Sindhu is the second Indian after Saina Nehwal to win in badminton after 2012"

